\documentclass{article}
\usepackage{spconf,amsmath,graphicx}
\usepackage{multirow}
\usepackage{booktabs}
\usepackage{amsfonts}
\usepackage{subfig}
\usepackage{stfloats}
\usepackage{CJKutf8}
\usepackage{xcolor}
\usepackage{amsmath}
\usepackage{adjustbox}
\usepackage{graphicx}
\usepackage{stackengine} 
\usepackage{mathtools}
\usepackage{hyperref}
\usepackage{url}

\newcommand\oast{\stackMath\mathbin{\stackinset{c}{0ex}{c}{0ex}{\ast}{\bigcirc}}}

\title{Good Neighbors Are All You Need for Chinese Grapheme-to-Phoneme Conversion}

\name{Jungjun Kim\textsuperscript{*}, Changjin Han\textsuperscript{*}, Gyuhyeon Nam, Gyeongsu Chae\textsuperscript{†}
\thanks{\textsuperscript{*}Equal contribution.
\textsuperscript{†}Correspondence to: \href{mailto:gc@deepbrain.io}{\texttt{gc@deepbrain.io}} \newline 
This work was supported by the Institute of Information \& Communications Technology Planning \& Evaluation (IITP) grant funded by the Ministry of Science and ICT (MSIT) of South Korea (No. 2021-0-00888).}}

\address{DeepBrain AI Inc., Seoul, Korea}

\begin{document}
\begin{CJK*}{UTF8}{gbsn}
\maketitle

\begin{abstract}
Most Chinese Grapheme-to-Phoneme (G2P) systems employ a three-stage framework that first transforms input sequences into character embeddings, obtains linguistic information using language models, and then predicts the phonemes based on global context about the entire input sequence. However, linguistic knowledge alone is often inadequate. Language models frequently encode overly general structures of a sentence and fail to cover specific cases needed to use phonetic knowledge. Also, a handcrafted post-processing system is needed to address the problems relevant to the tone of the characters. However, the system exhibits inconsistency in the segmentation of word boundaries which consequently degrades the performance of the G2P system. To address these issues, we propose the Reinforcer that provides strong inductive bias for language models by emphasizing the phonological information between neighboring characters to help disambiguate pronunciations. Experimental results show that the Reinforcer boosts the cutting-edge architectures by a large margin. We also combine the Reinforcer with a large-scale pre-trained model and demonstrate the validity of using neighboring context in knowledge transfer scenarios.

\end{abstract}
% %
\section{Introduction}
\label{sec:intro}

Grapheme-to-phoneme (G2P) conversion is a key procedure of the text-to-speech (TTS) system, which converts written-form texts into human-pronounced units called phonemes. G2P plays a crucial role in resolving the arbitrariness or irregularity of the relationship between written language and its spoken sound, contributing to the accurate pronunciation of synthesized speech. Chinese characters are ideograms representing ideas, not sounds, and pinyin is a system to map Chinese characters to their phoneme characters. The mapping is clear for monophonic characters, each corresponds to a single set of phoneme characters. However, it is challenging to assign correct phonemes to polyphonic characters because each polyphonic character has multiple pronunciations depending on its meaning within a sentence. Also, since different Chinese characters may share the same pronunciation, conversion to incorrect phonemes results in altered or nonsensical meanings. Therefore, polyphone disambiguation is essential to Chinese speech synthesis.

\begin{table}[t]
\centering
\scalebox{0.85}{
\begin{tabular}{lcccc}

\hline
\multicolumn{1}{c|}{Word length} & \multicolumn{1}{c}{\# of Word} & \multicolumn{1}{c}{\# of Polyphonic Word} \\ \hline\hline
\multicolumn{1}{l|}{1}     & 101,631,264 \textbf{(36.58\%)}                         & 30,538,738 \textbf{(36.38\%)}                                                                \\ 
\multicolumn{1}{l|}{2}     & 149,085,563 \textbf{(53.67\%)}                         & 42,038,705 \textbf{(50.08\%)}                                                               \\ 
\multicolumn{1}{l|}{3}     & 21,099,340 \textbf{(7.60\%)}                            & 8,275,774 \textbf{(9.86\%)}                                                              \\ 
\multicolumn{1}{l|}{$>$ 3} & 5,977,825 (2.15\%)                            & 3,094,409 (3.68\%)                                                               \\ \hline\hline

\multicolumn{1}{l|}{Total} & 277,793,992                             & 83,947,626                                                                 \\ \hline

\hline
\end{tabular}
}

\caption{Statistics of zhwiki\protect\footnotemark[1] dataset.}
\vspace{-0.5cm}

\label{table:zhwiki_stats}
\end{table}
\footnotetext[1]{\url{https://dumps.wikimedia.org/zhwiki/}}
Here, we propose the Reinforcer that learns the statistical dependencies of adjacent characters for polyphone disambiguation. In the relationship between a Chinese character and its sound, it is important to capture the frequency, regularity, and consistency of the neighboring characters in diverse sound-linguistic tasks \cite{glushko1979organization, sproat1990statistical, tsai2009maximum}. The Reinforcer emphasizes the local context between adjacent characters before a language model comes in. We suppose there are two aspects of this emphasis.

First, it reinforces the quasi-intra-word relationship of characters. The function and meaning of a word containing a polyphone determine the pronunciation of the polyphone. We analyze the word length frequency in a large Chinese corpus used to train the Chinese BERT \cite{cui2021pre}. Table 1 shows the most common word length is 2 (53.67\% of words; 50.08\% of polyphonic words) and the majorities are shorter than or equal to 3 (97.85\% of words; 96.32\% of polyphonic words). This observation supports that the neighboring characters have a significant impact on the pronunciation of Chinese polyphones. Then, as context expands from a word to a sentence, the expansion provides a contextual hierarchy to the language model. Second, it reinforces the phonological influence between neighboring characters (e.g., tone sandhi) regardless of whether they are in a word or not. Overall, the Reinforcer provides an effective way to incorporate hierarchical and phonological knowledges to language models.

\section{End-to-end Chinese G2P Conversion System}
\subsection{Base Setup}

Our G2P framework consists of three parts: Reinforcer, language model, and classifier. The input sequence of characters is denoted as $x=(x_{1}, ..., x_{T}) \in \mathbb{R}^{T}$. We maps sequence of characters $x$ into trainable real-valued feature vectors $E_{EMB} \in \mathbb{R}^{T \times D}$. The Reinforcer uses input sequence embedding $E_{EMB}$ to learn the associations between characters and implicitly obtain the meaning and function of character within a sentence for solving the polyphone disambiguation. The Reinforcer outputs contextualized character-level representations $E_{CXT} \in \mathbb{R}^{T \times D}$. The language model then absorbs the character representations for sequence modeling. Lastly, the classification layer is fed with the language model output $E_{LM} \in \mathbb{R}^{K \times D}$ and produces answer confidence ${A}^{*}$:

\begin{equation}
\begin{split}
&\boldsymbol{E}_{CXT} = Reinforcer(\boldsymbol{E}_{EMB}), \\
&\boldsymbol{E}_{LM} = LanguageModel(\boldsymbol{E}_{CXT}), \\
&\boldsymbol{A}^{*} = Classifier_{|dict|}(\boldsymbol{E}_{LM}),
\end{split}
\end{equation}

\noindent where $|dict|$ denotes the number of phoneme candidates in a dictionary, $K$ indicates the number of polyphones in a sentence. We determine whether Chinese graphemes are polyphones or monophones based on CC-CEDICT\footnote[2]{\url{https://cc-cedict.org/wiki/}}, a public-domain Mandarin Chinese dictionary that contains grapheme-phoneme mapping.

\subsection{Reinforcer: Neighborhood-based Character-level Representation}
In Mandarin Chinese, the pronunciations of Chinese polyphones are determined by the function or meaning of a word they belong to. Also, the phoneme changes depending on the tone of the adjacent characters known as tone sandhi. For example, {为} can be spoken as $wei4$ or $wei2$ according to adjacent characters or words. In Sentence 1, the phoneme of {为} following with {因} is $wei4$ when {为} has the meaning of ``$because$''. On the other hand, in Sentence 2, {为} is pronounced as phoneme $wei2$ meaning ``$as$'' when used together with {人处世}. Meanwhile, in Sentence 3, {首} is canonically pronounced by $shou3$. However, because two Chinese characters with tone 3 appear consecutively in one word, the tone of the preceding character is changed to $shou2$ by the tone sandhi rule.

\begin{itemize}
  \setlength\itemsep{0.2em}
  \item Sentence 1: 因为个人问题而请假 
  \item Phonemes: yin1 \textbf{wei4} ge4 ren2 wen4 ti2 er2 qing3 jia4
  \item Sentence 2: 为人处世方面还略有不足
  \item Phonemes : \textbf{wei2} ren2 chu3 shi4 fang1 mian4 hai2 lve4 you3 bu4 zu2
  \item Sentence 3: 首长的视察如期到来
  \item Phonemes : shou2 (Canonical: \textbf{shou3}) zhang3 de5 shi4 cha2 ru2 qi1 dao4 lai2
\end{itemize}

Therefore, contextual association relationships play an important role in pronouncing Chinese words. In the following subsections, we explore the simple but powerful neighboring-based character-level representation methods for the Chinese polyphone disambiguation task.

\subsubsection{1D Convolutional Neural Netwokrs}
The widely used method to extract character-level features is to use 1D convolutional neural networks \cite{dos2014learning, liang2017combining}. The convolutional neural networks with a specific kernel size are known to be effective in capturing morphological features about the character by sliding convolutional filters in all positions of a word. Meanwhile, we extend the range of the sliding window from within a word to within a sentence in order to learn more useful associations between characters.

The character sequence embeddings $\boldsymbol{E}_{EMB}$ is fed to a 1D convolutional layer. As a result, the convolutional layer produces the sequence of vectors $E_{CXT}$:

\begin{equation}
\begin{split}
&\boldsymbol{E}_{CXT} = Conv(\boldsymbol{E}_{EMB}), \\
&Conv(X) \vcentcolon= W \oast X,
\end{split}
\end{equation}

\noindent  where $E_{EMB} \in \mathbb{R}^{T \times D}, E_{CXT} \in \mathbb{R}^{T \times D}$, $W$ means the learnable weights of convolutional filters, and $\oast$ denotes 1D convolution operation. We set the padding size and stride as 1 to retain the input sequence length. The biases are omitted to simplify notations. We call the 1D convolutional layer ``Conv'' for convenience.

\subsubsection{Shift and Stack Operation}

In recent computer vision, \cite{tolstikhin2021mlp, liu2021pay, lian2022asmlp, yu2022s2} proposed various MLP-based patch-wise operations instead of convolutional operations. MLP-Mixer \cite{tolstikhin2021mlp} has firstly explored the potential of MLP for image classification as a pioneer. MLP-Mixer introduced channel-mixing and token-mixing techniques to capture channel-wise relationships and communicate spatial locations. Unlike MLP-Mixer, AS-MLP \cite{lian2022asmlp} performs the axial shift that successively shifts the patch-tokens along the vertical and horizontal axis. The axial shift block focuses on local feature communications allowing the information to flow within the receptive field. As the axial shift block is layered, the receptive field is gradually enlarged such that the blocked tokens information can be distilled into another one. Inspired by that, we introduce a simply modified shift and stack operation (SSO) for character-level representation in the Chinese G2P task. AS-MLP axially shifts the channels by slicing the input features, whereas SSO treats tokens as individual units and shifts and stacks the entire token embedding sequence. This enables SSO to fully utilize the contextual information without repeating blocks of the same operation. To be specific, given the character sequence embeddings $\boldsymbol{E}_{EMB}$, SSO conducts as follows: 

\begin{figure}[]
\includegraphics[width=\columnwidth]{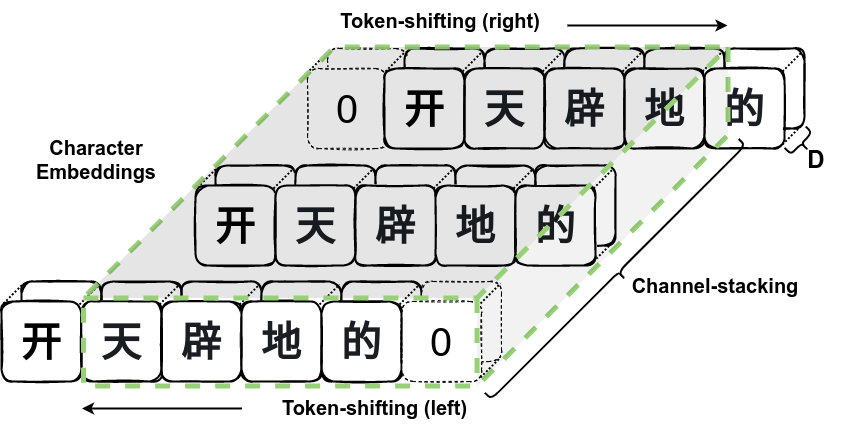}
\caption{Illustration of the shift and stack operation.}
\label{figure:SSO}
\end{figure}

\begin{equation}
\setlength{\jot}{7pt}
\renewcommand\theequation{3}
\begin{split}
&\boldsymbol{E}^{k\in \mathbb{Z}}_{EMB} \leftarrow \textbf{0}[0:T,0:D],\\
&\boldsymbol{E}^{k<0}_{EMB}[0:T+k, 0:D] \leftarrow \boldsymbol{E}_{EMB}[-k:T, 0:D], \\
&\boldsymbol{E}^{k=0}_{EMB}[0:T, 0:D] \leftarrow \boldsymbol{E}_{EMB}[0:T, 0:D], \\
&\boldsymbol{E}^{k>0}_{EMB}[k:T, 0:D] \leftarrow \boldsymbol{E}_{EMB}[0:T-k, 0:D], \\
&\boldsymbol{\hat{E}}^{w}_{EMB} = Concat(\{\boldsymbol{E}^{k}_{EMB}\}^{s}_{k=-s}), w=2s+1, \\
&\boldsymbol{E}_{CXT} = \sigma(\boldsymbol{\hat{E}}^{w}_{EMB}\bar{\boldsymbol{W}})+\boldsymbol{E}_{EMB},
\end{split}
\end{equation}
 
\noindent where $\boldsymbol{0}$ means the $D$-dimensional zero-padded vector, $s$ indicates shift size, $w$ is window size, and $T$ is sequence length. $\sigma$ is GELU activation function \cite{hendrycks2016gaussian}. $\bar{\boldsymbol{W}} \in \mathbb{R}^{(w \times D) \times D}$ is learnable parameters of linear projection layers.

\section{Experiments and Results} \label{sec:exp}
\subsection{Experiment details}
In our experiments, we take Transformer \cite{vaswani2017attention} and MLP-Mixer \cite{tolstikhin2021mlp} as baseline architectures and also employ BERT \cite{devlin-etal-2019-bert} to confirm the improvement in a model trained with giant corpus. We vary the number of layers to observe performance gain according to the model capacity. Note that the layer refers to the encoder layer of Transformer and the mixer layer of MLP-Mixer. We conduct experiments on the DataBaker dataset\def\thefootnote{3}\footnote{\url{https://en.data-baker.com/\#/data/index/source}}, an open-source benchmark for polyphone disambiguation. The dataset consists of Mandarin Chinese grapheme-phoneme pairs of 10,000 sentences. We use 9,776 samples after removing the sentences with unequal source and target lengths. The total dataset contains 4,112 unique Chinese graphemes and 1,536 unique phonemes, in which 27\% are polyphones.

We train each model using Adam optimizer with a learning rate of 1e-4 and a batch size of 256. The character embedding size is 256 except for BERT which is 768. We use the cross-entropy loss with label smoothing of 0.1 to optimize our models. We use the kernel size of 3 for Conv. We also set the window size to 3 with a shift size of 1 for SSO. We randomly divide the dataset into training, validation, and test sets by a split ratio of 8:1:1. In order to minimize the effect of data split, we iterate the random splitting three times with different seeds and average the three results.

\subsection{Plugging the Reinforcer into baseline architectures}
Table \ref{table:performance} presents polyphone disambiguation accuracy of baseline models (i.e., Transformer and MLP-Mixer) and the Reinforcer-combined models (i.e., Conv-Transformer, SSO-Transformer, Conv-MLP-Mixer, and SSO-MLP-Mixer) according to the number of layers. As shown in Table \ref{table:performance}, the Reinforcer-combined models outperform the baseline models by margins of 5.79\% and 2.68\% for Transformer and MLP-Mixer, respectively. The Conv-MLP-Mixer(Layer=2) achieves the highest accuracy. In Transformer, the effect size of Conv and SSO is similar, but Conv is more powerful than SSO in MLP-Mixer. Since Conv and SSO have same parameter size, the result can be interpreted that the gap is originated from the difference of operation to deal with channel information.

\begin{table}[]
\resizebox{\columnwidth}{!}{
\begin{tabular}{lcccc}
\toprule
\multicolumn{1}{l|}{Model}                       & \multicolumn{1}{c|}{$\# of Layers$ = 1}         & \multicolumn{1}{c|}{$\# of Layers$ = 2}                  & \multicolumn{1}{c|}{$\# of Layers$ = 4}              & $\# of Layers$ = 8         \\ \midrule\midrule
\multicolumn{1}{l|}{Transformer}            & \multicolumn{1}{c|}{86.75$\pm${1.03} (2.5M)} & \multicolumn{1}{c|}{87.21$\pm${1.68} (3.5M)} & \multicolumn{1}{c|}{86.72$\pm${1.22} (5.6M)}      & 86.42$\pm${1.25} (9.8M) \\
\multicolumn{1}{l|}{Conv-Transformer}       & \multicolumn{1}{c|}{92.56$\pm${1.15} (2.7M)}     & \multicolumn{1}{c|}{92.85$\pm${1.34} (3.7M)}     & \multicolumn{1}{c|}{\textbf{92.82}$\pm${1.31} (5.8M)}          & \textbf{92.61}$\pm${1.23} (10M)  \\
\multicolumn{1}{l|}{SSO-Transformer}        & \multicolumn{1}{c|}{\textbf{92.96}$\pm${1.12} (2.7M)}     & \multicolumn{1}{c|}{\underline{\textbf{93.00}}$\pm${1.22} (3.7M)}     & \multicolumn{1}{c|}{92.59$\pm${1.24} (5.8M)}          & 92.48$\pm${1.44} (10M)     \\ \midrule\midrule
\multicolumn{1}{l|}{MLP-Mixer}              & \multicolumn{1}{c|}{90.86$\pm${1.04} (1.5M)}     & \multicolumn{1}{c|}{91.04$\pm${1.03} (1.7M)}              & \multicolumn{1}{c|}{91.10$\pm${1.08} (2.0M)} & 90.67$\pm${0.79} (2.5M)    \\
\multicolumn{1}{l|}{Conv-MLP-Mixer}         & \multicolumn{1}{c|}{\textbf{93.63}$\pm${0.86} (1.7M)}     & \multicolumn{1}{c|}{\underline{\textbf{93.72}}$\pm${0.91} (1.9M)}     & \multicolumn{1}{c|}{\textbf{93.68}$\pm${0.87} (2.2M)}          & \textbf{93.58}$\pm${0.74} (2.7M)    \\
\multicolumn{1}{l|}{SSO-MLP-Mixer}          & \multicolumn{1}{c|}{93.22$\pm${0.91} (1.7M)}     & \multicolumn{1}{c|}{93.28$\pm${1.12} (1.9M)}              & \multicolumn{1}{c|}{93.50$\pm${0.94} (2.2M)} & 93.45$\pm${0.96} (2.7M)    \\
 \bottomrule
\end{tabular}}
\caption{Polyphone accuracy on the test split of DataBaker.}
\label{table:performance}
\end{table}

\begin{table}[]
\centering
\scalebox{0.8}{
\begin{tabular}{lllll} \hline
\multicolumn{1}{|l|}{Model}         & \multicolumn{1}{l|}{BERT}       & \multicolumn{1}{l|}{Conv-BERT}   & \multicolumn{1}{l|}{SSO-BERT}    \\ \hline
\multicolumn{1}{|l|}{Poly Acc (\%)} & \multicolumn{1}{l|}{88.48$\pm$1.65} & \multicolumn{1}{l|}{92.37$\pm$1.24} & \multicolumn{1}{l|}{92.82$\pm$1.13}   \\ \hline
\end{tabular}
}
\caption{Performance of BERT with the Reinforcer.}
\label{table:bert}
\end{table}

\subsection{Plugging the Reinforcer into BERT}
\begin{figure*}[t]
\includegraphics[width=\textwidth]{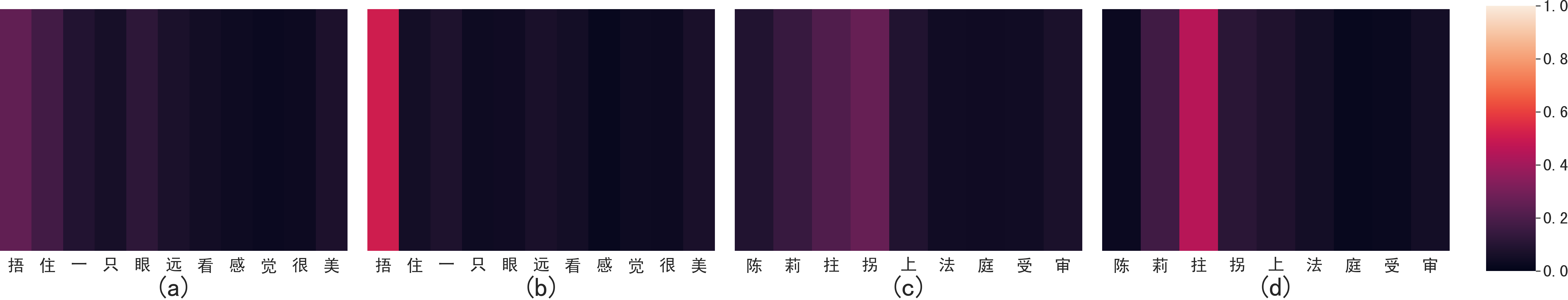}
\caption{Heatmaps of attention weights averaged across all layers and heads of BERT and SSO-BERT. (a) and (b) show attention weights of BERT and SSO-BERT, respectively, to determine the phoneme of ``捂'' in sentence ``捂住一只眼远看感觉很美''. Likewise, (c) and (d) also display attention weights of the both models for ``拄'' in sentence ``陈莉拄拐上法庭受审''.}
\label{figure:attention_heatmap}
\end{figure*}

To see the impact of the Reinforcer on pre-trained model, we use Chinese BERT\def\thefootnote{4}\footnote{\url{https://huggingface.co/bert-base-chinese}}. Following Dai et al. \cite{dai2019disambiguation}, we freeze the parameters of BERT to avoid the degradation due to over-fitting and train only the Reinforcer and classifier. Table \ref{table:bert} summarizes the performance comparisons according to the type of the Reinforcer. Conv-BERT and SSO-BERT show 3.89\% and 4.34\% improvements compared to BERT, respectively. Since Chinese characters can have the same meaning but different pronunciations, we reason BERT embeddings that focus more on meaning lead to relatively low performance. We carefully assume that the Reinforcer not only provides prior semantic knowledge, but also makes the phonetic information richer for language models by capturing context from neighboring characters. To take a closer look at the Reinforcer's effect, we extract attention weights for a polyphone of BERT and SSO-BERT when only the prediction of SSO-BERT is correct, and visualize as attention heatmaps. As shown in Figure \ref{figure:attention_heatmap}, SSO-BERT highly concentrates on the target polyphone's representation containing neighboring context, while BERT imposes a similar level of weights around the polyphone. Thus, we argue that the representations obtained by the Reinforcer contain rich phonetic information to solve polyphone disambiguation.

\subsection{Case Study}
We study enhanced cases due to the Reinforcer compared to MLP-Mixer which is baseline. In Case 1 of Table \ref{table:case_study}, the tone of {一} is changed to 2 when a letter with tone 4 is followed, otherwise, it has a tone 4 canonically. In addition, Case 2 and Case 3 deal with the tone sandhi rule where {有} and {两} in parentheses of sentence originally have the pronunciation of $you3$ and $liang3$, respectively. According to the rule, both are affected by the following phonemes with tone 3 and changed to tone 2. Since the Reinforcer+MLP-Mixer corrects phoneme considering the tone sandhi rule, we speculate that character representations implicitly learn the phonetic information of Chinese character by augmenting the neighboring context. Meanwhile, we note that the handcrafted tone change system is challenging because applying the rule relies on word segment tools that can cause errors. In Case 2, the tone of {有} is not changed to 2 when using Jieba\def\thefootnote{5}\footnote{\url{https://github.com/fxsjy/jieba}} since it is included in a four-letter word. On the other side, in Case 3, Pkuseg\def\thefootnote{6}\footnote{\url{https://github.com/lancopku/pkuseg-python}} separates {两} from two-letter word {两种} and the tone sandhi rule is not applicable. As a result, the manual system arbitrarily operates depending on the external word segmentors. However, the Reinforcer is advantageous for obtaining the quasi-intra-word concept of the character without external word segment tools.

\begin{table}[ht]
\centering
\resizebox{\columnwidth}{!}{
\begin{tabular}{l|l|ll}
\toprule
\multirow{5}{*}{Case 1} & Sentence  & \multicolumn{2}{l}{\begin{tabular}[c]{@{}l@{}}一年(一)度的高考 \end{tabular}}                                                                                 \\ \cmidrule{2-4} 
                        & Phonemes   & \multicolumn{2}{l}{\begin{tabular}[c]{@{}l@{}}Target: yi4 nian2 (\textbf{yi2}) du4 de5 gao1 kao3 \\ Pred: yi4(MLP-Mixer) / \textbf{yi2}(Reinforcer+MLP-Mixer)\end{tabular}}                                                 \\ \cmidrule{2-4} 
                        & Segments & \multicolumn{2}{l}{\begin{tabular}[c]{@{}l@{}}Jieba: (一年\underline{一}度) (的) (高考) \\ Pkuseg: (一年\underline{一}度) (的) (高考) \end{tabular}}                                                \\ \midrule
\multirow{6}{*}{Case 2} & Sentence  & \multicolumn{2}{l}{\begin{tabular}[c]{@{}l@{}}跟我们现在的年代是(有)所区别的 \end{tabular}}                                                                         \\ \cmidrule{2-4} 
                        & Phonemes   & \multicolumn{2}{l}{\begin{tabular}[c]{@{}l@{}}Target: gen1 wo3 men5 xian4 zai4 de5 nian2 dai4 shi4 \\(\textbf{you2}) suo3 qu1 bie2 de5 \\ Pred: you3(MLP-Mixer) / \textbf{you2}(Reinforcer+MLP-Mixer) \end{tabular}}\\ \cmidrule{2-4} 
                        & Segments & \multicolumn{2}{l}{\begin{tabular}[c]{@{}l@{}}Jieba: (跟) (我们) (现在) (的) (年代) (是) (\underline{有}所区别) (的) \\ Pkuseg: (跟) (我们) (现在) (的) (年代) (是) (\underline{有}所) (区别) (的)\end{tabular}}                          \\ \midrule
\multirow{5}{*}{Case 3} & Sentence  & \multicolumn{2}{l}{\begin{tabular}[c]{@{}l@{}}找出(两)种填在这里\end{tabular}}                                                                           \\ \cmidrule{2-4} 
                        & Phonemes   & \multicolumn{2}{l}{\begin{tabular}[c]{@{}l@{}}Target: zhao3 chu1 (\textbf{liang2}) zhong3 tian2 zai4 zhe4 li3 \\ Pred: liang3(MLP-Mixer) / \textbf{liang2}(Reinforcer+MLP-Mixer) \end{tabular}}                                                                                                      \\ \cmidrule{2-4} 
                        & Segments & \multicolumn{2}{l}{\begin{tabular}[c]{@{}l@{}}Jieba: (找出) (\underline{两}种) (填) (在) (这里)\\ Pkuseg: (找出) (\underline{两}) (种) (填) (在) (这里)\end{tabular}}                                                 \\ \bottomrule
\end{tabular}}
\caption{Examples used in case study.}
\label{table:case_study}
\end{table}
\vspace{-0.5cm}
\section{Discussion}
\label{sec:discussion}

This paper introduces a novel approach to reinforce character representation in Chinese G2P. The proposed approach can be easily plugged into existing neural architectures, such that boosts the performance of prediction accuracy. Also, this approach adds the phonological information of left-and-right sides to a character representation and contributes to solving tone-relevant problems. Experimental results on a public Mandarin dataset show that the Reinforcer improves the conversion quality of Transformer, MLP-Mixer, and BERT. We visualize the attention heatmap of BERT and observe that BERT’s character embedding itself combined with the Reinforcer have the highest weights. The case study on Chinese tone supports the Reinforcer efficiently encodes the tonal knowledge of neighboring characters.

\bibliographystyle{IEEEbib}
\bibliography{mybib}
\end{CJK*}
\end{document}